\documentclass[12pt]{article}
\usepackage{graphicx}%
\usepackage{multirow}%

\usepackage{amsmath,amssymb,amsfonts}%
\usepackage{amsthm}%
\usepackage{mathrsfs}%
\usepackage{url} 
\usepackage{xcolor}%
\usepackage{textcomp}%
\usepackage{manyfoot}%
\usepackage{booktabs}%
\usepackage{algorithm}%
\usepackage{algorithmicx}%
\usepackage{algpseudocode}%
\usepackage{listings}%
\usepackage{adjustbox}
\usepackage{indentfirst}
\usepackage{authblk}  
\usepackage{amsmath,amssymb}  
\usepackage{graphicx}          
\usepackage{geometry}          
\begin{document}

\title{\textbf{\Large MicroRoboScope: A Portable and Integrated Mechatronic Platform for Magnetic and Acoustic Microrobotic Experimentation}}

\author[1]{Max Sokolich}
\author[1]{Yanda Yang}
\author[1]{Subrahmanyam Cherukumilli}
\author[2]{Fatma Ceren Kirmizitas}
\author[1,*]{Sambeeta Das}

\affil[1]{Department of Mechanical Engineering, University of Delaware, 130 Academy St, Newark, Delaware 19716, USA}
\affil[2]{Departments of Animal \& Food Sciences, Biological Sciences, and Medical \& Molecular Sciences, University of Delaware, Newark, DE 19716, USA}

\affil[*]{Corresponding author: \texttt{samdas@udel.edu}}
\date{}

\maketitle

\abstract{This paper presents MicroRoboScope, a portable, compact, and versatile microrobotic experimentation platform designed for real-time, closed-loop control of both magnetic and acoustic microrobots. The system integrates an embedded computer, microscope, power supplies, and control circuitry into a single, low-cost and fully integrated apparatus. Custom control software developed in Python and Arduino C++ handles live video acquisition, microrobot tracking, and generation of control signals for electromagnetic coils and acoustic transducers. The platform’s multi-modal actuation, accessibility, and portability make it suitable not only for specialized research laboratories but also for educational and outreach settings. By lowering the barrier to entry for microrobotic experimentation, this system enables new opportunities for research, education, and translational applications in biomedicine, tissue engineering, and robotics.}

\section{Introduction}
Microscale robots have a variety of potential applications in medicine, environmental monitoring, and tissue engineering, due to their small size and capabilities of sensing and manipulation at the small scale \cite{doi:10.1021/acsnano.3c03723}. Recent research has demonstrated their potential in applications ranging from ocular drug delivery and in vitro fertilization to root canal prevention and tumor treatment \cite{li2025inhalable,tran2025robotic}. The most common actuation methods for microscale robots are acoustic and electromagnetic actuation \cite{chen2022electromagnetic}. Acoustic microrobots, for instance, can be manipulated using sound waves to achieve precise movements, while electromagnetic microrobots rely on magnetic fields for their actuation and control. Traditional open-loop control systems for acoustic and magnetic microrobots often fail to provide the necessary accuracy and reliability required for the above applications \cite{shen2023magnetically}. Therefore, it is necessary to integrate closed loop control with these actuation systems to significantly enhance their accuracy and performance. Naik et al. have demonstrated control using an acoustic phase array for manipulating micro objects in air \cite{10440624}, however, they don't use magnetic fields and the approach operates in open loop rather than closed loop.  Additionally, in \cite{8629946}, Youssefi et al developed a magneto acoustic system for closed loop manipulation of objects in air, however, the apparatus is not portable and constrained to a desktop computer. Additionally, the system handles microrobots on the order of millimeters. 

As researchers seek to bridge the gap between laboratory prototypes and real-world applications, compact and portable experimental systems are needed to quickly test the efficacy of microrobots. We have previously developed the ModMag \cite{sokolich2023modmag} which is a low-cost portable magnetic microrobot manipulation device. However, this system has no feedback feature and applies magnetic and acoustic signals entirely in open loop. Other systems like MicrostressBots have also been developed \cite{foroutan2018sat}. However, these systems often require specialized equipment and are not easily transportable, limiting their use in field studies. 

\begin{figure}
    \centering
    \includegraphics[width=.8\linewidth]{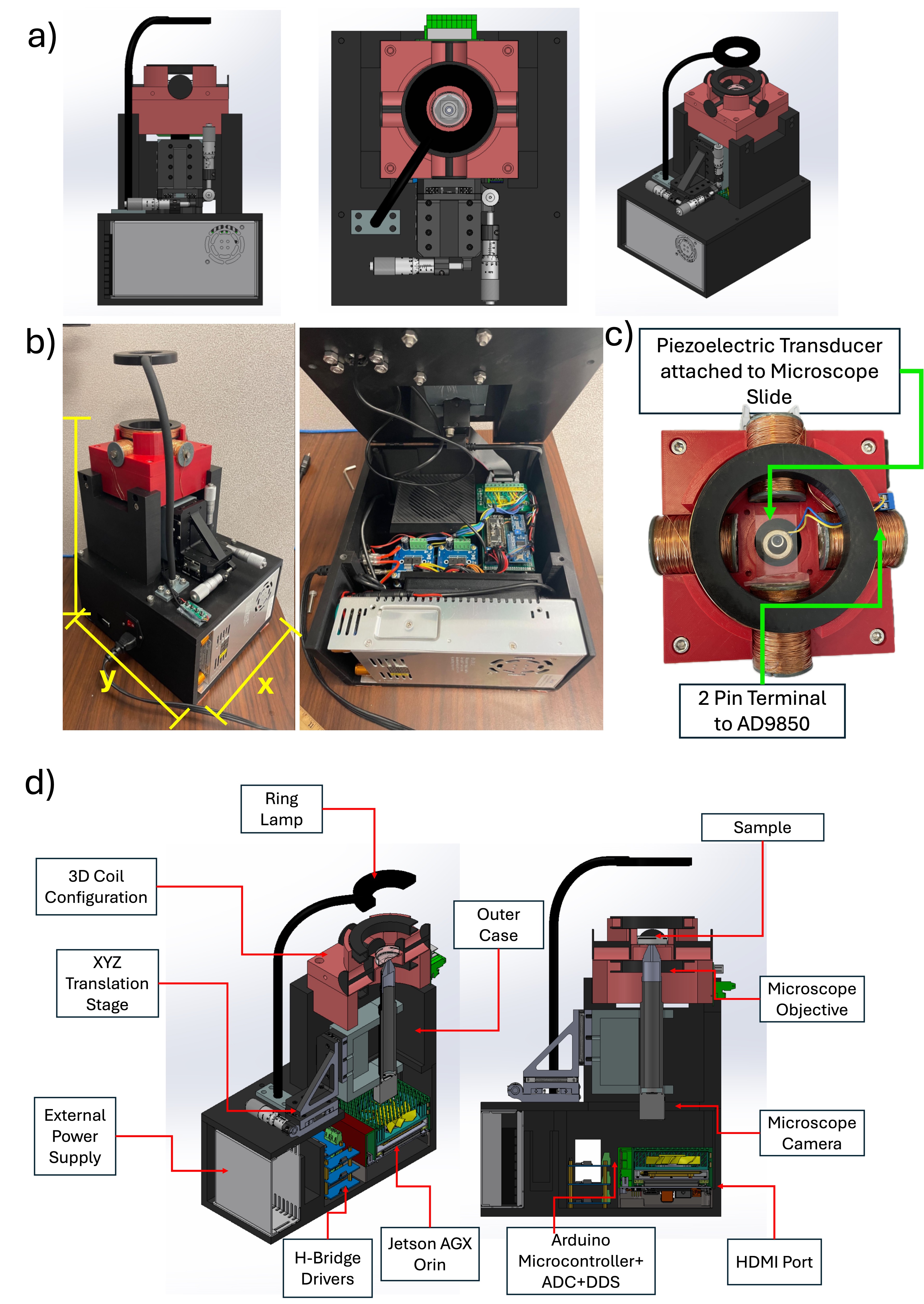}
    \caption{a) Back, side, front, top and isometric views of computer aided design of device.  b) Images of the physical system. The system measures 240 mm in the x-direction, 280 mm in the y direction, and 350 mm in the z-direction. These dimensions allow for all custom-made components, including the outer case, to be 3D printed on an Ender Max 3D printer. All electronics are housed inside the outer-case and accessible. c) Coil configuration with piezoelectric transducer and connections labeled. d) Computer aided design of the cross section of the device with important system components highlighted and defined.}
    \label{fig:Figure1}
\end{figure}

This paper introduces the development of a compact, portable closed-loop microrobotic experimentation platform, which can combine acoustic and electromagnetic actuation (see Figure \ref{fig:Figure1}). The proposed platform integrates real-time visual feedback mechanisms to continuously monitor and adjust the position and velocity of microrobots. The novelty of the system arises from its significant integration of fundamental microrobotic functionality in a compact and portable manner. By leveraging acoustic and electromagnetic control techniques, this platform aims to overcome the limitations of traditional open-loop systems and provide a robust solution for various microrobotic applications. 

\section{Background}
\subsection{Magnetic Actuation}
Microrobots are typically actuated via external magnetic fields. By applying an external magnetic field on a magnetic microrobot, the microrobot's magnetic moment \textbf{m} will want to align with the direction of the applied field. The subsequent torque, $\boldsymbol{\Gamma}$, is given by
\begin{equation}
    \boldsymbol{\Gamma} = \textbf{m}\times\textbf{B}
\end{equation}
\noindent where $\textbf{m}$ is the magnetic moment of the microrobot and $\textbf{B}$ is the magnetic flux density \cite{5595508}. This torque therefore allows one to magnetically orient the microrobot in 3D space.  

Magnetic forces can also be applied to microrobots by applying magnetic gradients. The magnetic force on a microrobot, $\textbf{F}$, is given by
\begin{equation}
    \textbf{F} =  \left(\textbf{m}\cdot\nabla\right)\textbf{B}
\end{equation}
where $\textbf{m}$ is the magnetic moment of the microrobot and $\textbf{B}$ is the magnetic flux density \cite{5595508}. The magnetic forces need to be sufficiently strong to overcome the viscous forces that dominate behavior at the microscale. In practice, it is often difficult to produce adequately strong magnetic forces on microrobots unless the workspace is very small.  As a result, it is often advantageous to utilize rotating magnetic fields. These fields result in rolling, spinning, or swimming motion of magnetic microrobots rather than pulling, dragging, or translational motion.

\subsection{Acoustic Actuation}
Microrobots can also be actuated using external acoustic fields \cite{aghakhani2020acoustically}. An acoustic bubble-propelled microrobot typically contains a cavity or hole in its design. When placed in a fluid environment, a gas bubble is trapped in the cavity of the microrobot. The gas bubble is excited via a piezoelectric transducer which generates bubble propulsion. The microrobot is subjected to various forces under acoustic fields such as acoustic radiation and streaming forces \cite{bertin2015propulsion}. The acoustic radiation forces include the Primary Bjerknes and Secondary Bjerknes forces \cite{ren20193d}. The Primary Bjerknes forces are due to the scattering of the acoustic waves by the oscillating bubble. These are in the same direction as the direction of wave propagation. These forces are negligible due to the attenuation of the waves being much larger than the microrobot size. The two main forces that significantly impact the microrobot motion are the streaming propulsive force and the secondary Bjerknes force \cite{ahmed2016artificial,bertin2015propulsion, ren20193d}. The Secondary Bjerknes force creates an attraction of the microrobot to the underlying solid surface. The presence of the boundary results in a boundary condition that is satisfied by considering an image bubble that oscillates in sync with the real bubble, creating an attractive force between the two bubbles \cite{ren20193d}. The acoustic streaming forces are generated by the pulsating bubble in the cavity which creates fluid flows that propel the microrobot away from the open end. These forces are maximized at the resonant frequency of the bubble. 

The resonant frequency at which the microrobot moves with its highest velocity is theoretically described via \eqref{eq:freq} and \eqref{eq:M}, as follows:

\begin{equation}
    f_0 = \frac{1}{2\pi}\Bigg(\frac{\kappa P_0}{\rho(L - L_b)L_b}\Bigg)^{1/2}\times M,
    \label{eq:freq}
\end{equation}

\begin{equation}
    M=\Bigg(1+\frac{4\gamma L_b}{\kappa P_0 a^2}\Bigg)^{1/2},
    \label{eq:M}
\end{equation}
where $\kappa \sim 1.4$ is the adiabatic index, $\rho$ is the density of water, $P_0$ is the pressure in the bubble without an acoustic field, $L_b$ is the length of the bubble, $L$ is the cavity length, $a$ is the inner radius of the cavity, and $\gamma \sim 0.07 $\,N/m is the surface tension of the water-air interface \cite{ren20193d}. 

By regulating the frequency of the transducer, we gain control over the microrobot's velocity by moving into or out of resonance. This approach provides control over acoustic actuation, which is a versatile way to manipulate the microrobot.

\begin{figure}
    \centering
    \includegraphics[width=.9\linewidth]{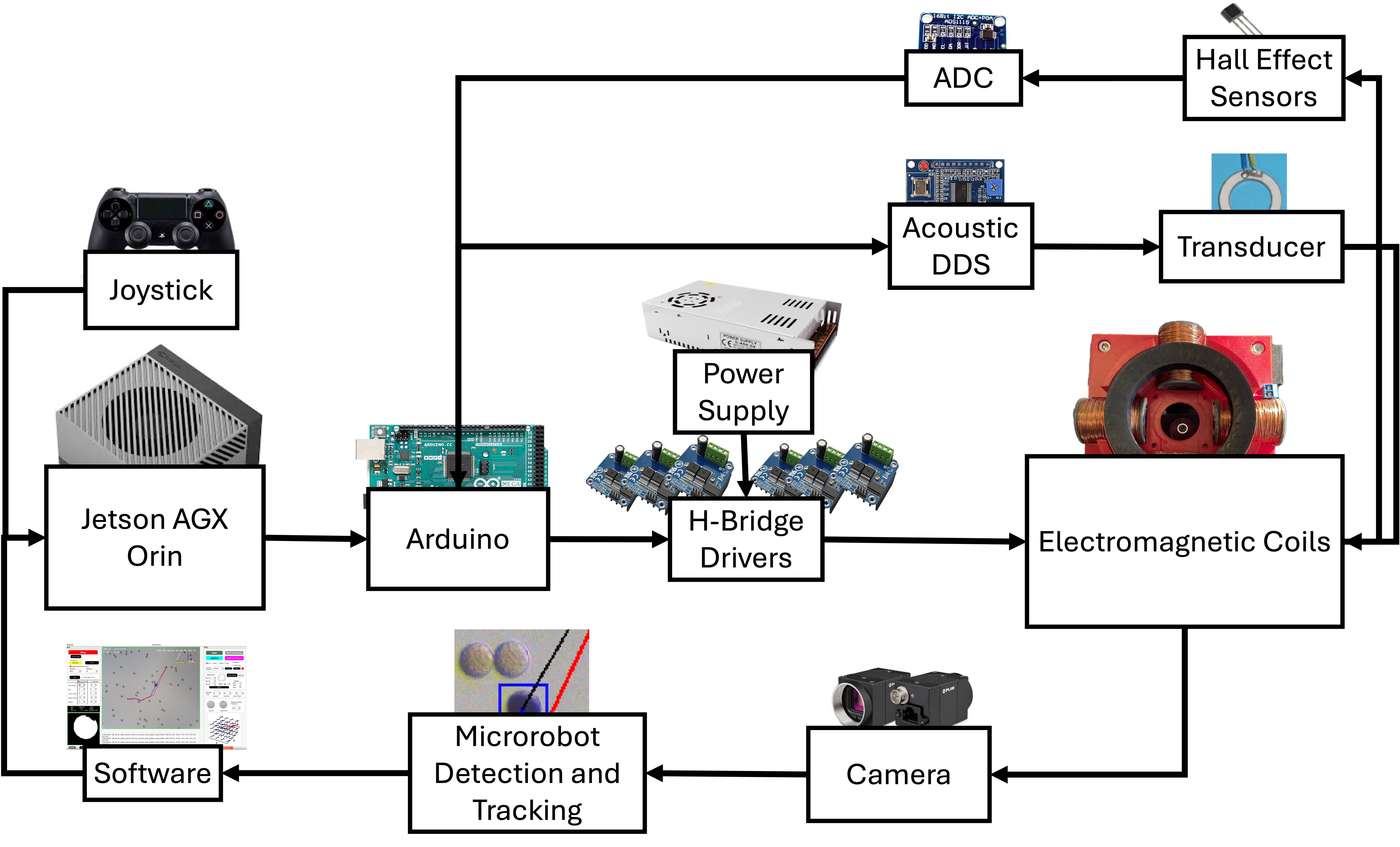}
    \caption{Flowchart of electrical system components. The Jetson AGX Orin communicates with an Arduino over serial to generate appropriate PWM (Pulse Width Modulation) signals. H-bridge circuits connected to an external power supply are used to switch the low current PWM signals from the arduino into high-current signals need to power the coils.  The Arduino also interfaces with an ADS1115 analog to digital converter to read hall effect sensor data, as well as an AD9850 signal generator module to output high frequency sine waves needed to drive an piezoelectric transducer. Image feed from a microscope camera captures the workspace environment and used for tracking and detection of microrobots.}
    \label{fig:Figure2}
\end{figure}

\section{Results}
First, we outline the mechanical, electrical, and software specifications of the system. Then we describe the method of magnetic field generation. Then, we showcase measurements of the magnetic field generated by the system. Then, three microrobotic actuation experiments were conducted to demonstrate the versatility of the system for microrobotic experimentation.  We demonstrate the system's ability to generate a rotating field that results in the autonomous translational rolling motion of a magnetic microrobot along a circular path. We also demonstrate the ability to control the rotating field and therefore the rolling microrobot manually with a joystick controller. We show precise manipulation by pushing non-magnetic passive microspheres while simultaneously tracking the position and velocity.  Finally, we demonstrate the systems ability to generate both magnetic fields and acoustic fields simultaneously closed loop. In this experiment, a 3 um cup-shaped acoustic microrobot is autonomously guided to a target coordinate in the workspace by regulating the acoustic frequency, microrobots velocity and magnetic field orientation direction in real time. Finally, we end with a conclusion.

\subsection{System Specifications}
\subsubsection{Mechanical}
The system measures 240 mm in the x-direction, 280 mm in the  y-direction, and 350 mm in the z-direction. The 3D coil configuration consists of 4 standard solenoids arranged about the XY plane, and a pair of Helmholtz coils in the Z direction. This coil configuration allows for 3D magnetic fields to be generated in the workspace. The specifications of these coils are as follows. The core of the standard solenoids are made of electrical steel with a diameter of 8\,mm and a length of 45\,mm. The relative magnetic permeability of the core is approximately 4000. The outer diameter of the entire coil measures 35\,mm resulting in 1500 turns of 24\,AWG copper wire. Opposite facing solenoids have a separation distance of 50\,mm allowing a standard 35\,mm petri dish to sit comfortably at the center, or half of a standard 75\,mm by 25\,mm glass microscope slide.  The pair of Helmholtz coils have an inner diameter of 70\,mm and an outer diameter of 100\,mm. The thickness of the coil is 10\,mm and there are 400 turns of 24\,AWG copper wire. The two coils are placed on the top and bottom of the 4 standard coils resulting in a coil separation distance of 35\,mm. All 6 coils are mounted on a custom stand designed in Solidworks and printed using an Ender 3 Max 3D printer. The 3D coil configuration is then mounted onto the outer case of the device using standard M6 bolts.

The Microscope aspect of the device includes a LX30/M - Self-Contained XYZ 25 mm Translation Stage mounted to a FLIR BFS-U3-50S5C-C USB 3.1 Blackfly® S, Color Camera.  This allows for precise focusing of the substrate using the Z axis micrometer, and translational scanning of the workspace scanning using the X and Y axis micrometers. To mount the camera to the stage, a custom adapter is designed and 3D printed. Then, C-mount extension tubes are used to establish an adequate optical tube length between the camera sensor and a 10x objective lens. The XYZ translation stage is mounted to the outer case below the 3D coil configuration.

The entire outer case is split in half and attached with hinges allowing for convenient inspection of the electric components when necessary (see Figure \ref{fig:Figure1}b). Finally, an adjustable ring lamp is attached to the outer case to illuminate the sample. The ring lamp's brightness can be adjusted along with options for white, warm white, and warm yellow light output. 

\begin{figure}
    \centering
    \includegraphics[width=\textwidth]{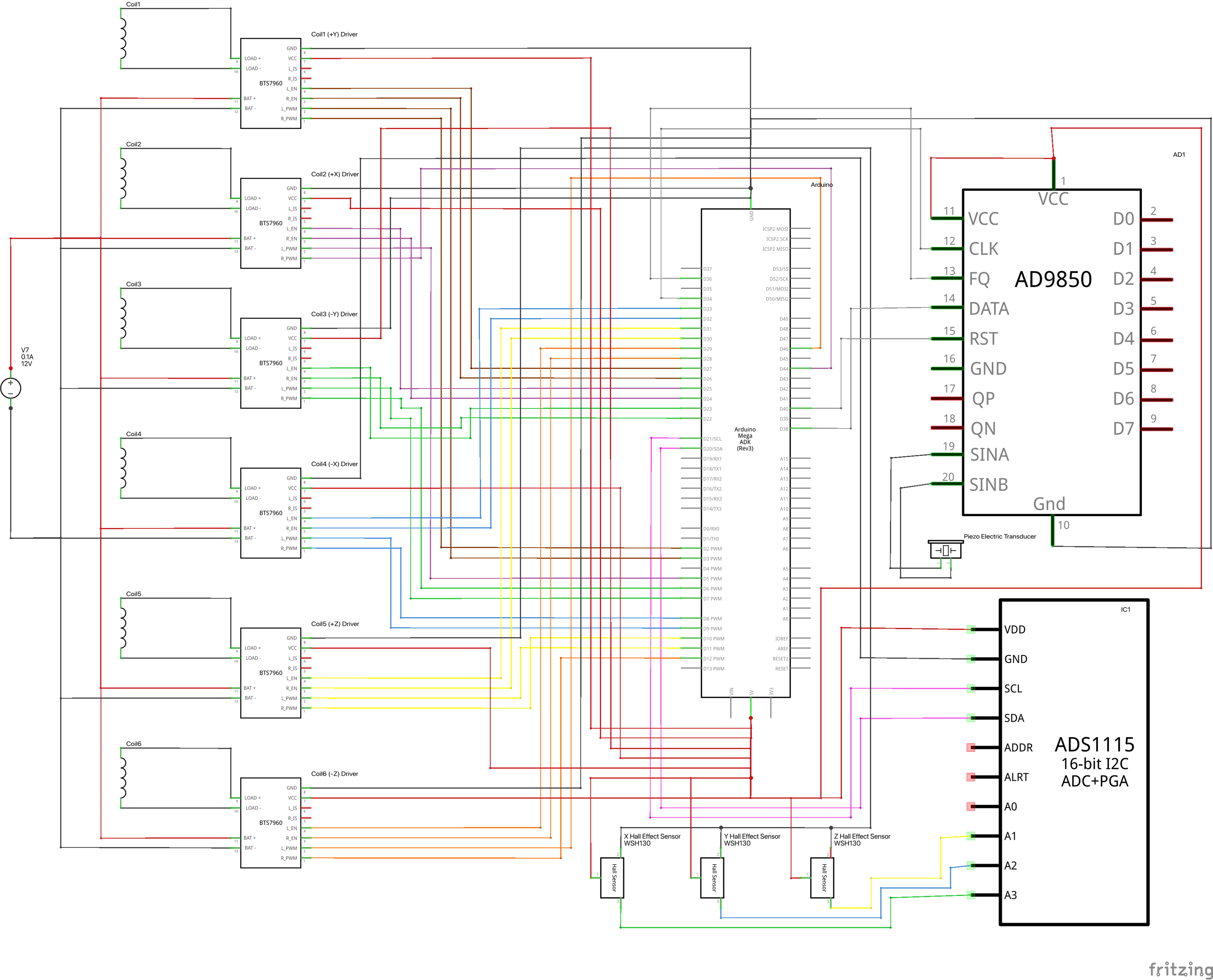}
    \caption{
        Arduino Electrical Schematic of the System.}
    \label{fig:Figure3}
\end{figure}

\subsubsection{Electrical}
The outer case houses the electrical components required for device operation. To begin, a Nvidia Jetson AGX Orin acts as the host computer for the system. The Jetson Orin single board computer enables the use of powerful parallel computing and GPU functionality which can enable more advanced microrobotic machine learning applications in future work. A FLIR BFS-U3-50S5C-C USB 3.1 Blackfly® S, Color Camera is connected to the computer via USB. In order to separate the output control signal thread from the camera processing thread, an Arduino Mega micro-controller is connected to the Jetson Orin via USB. The Arduino is connected to six SEEU. AGAIN BTS7960B 43A motor H-Bridge PWM drivers, one for each of the 6 coils. These are used to modulate the large currents necessary to power the electromagnetic coils, as well as switch the polarity of the field. An external 0-27\,V, 0-20\,A adjustable power supply is used to supply the necessary power to the coils.  The external power supply is connected to the H-Bridge drivers. To output the high frequency sine waves necessary to actuate a piezoelectric transducer and thus an acoustic microrobot, a HiLetgo DDS AD9850 Signal Generator Module is connected to the Arduino. This module is capable of generating sinusoidal waveforms from 0 to 40\,MHz at approximately 1\,VPP and square waves from 0 to 1\,MHz at approximately 1\,VPP. These frequency ranges are more than adequate for experimenting with a wide variety of acoustic microrobots. Furthermore, an ADS1115 16-Bit 16-Byte 4-Channel I2C IIC Analog-to-Digital ADC is connected to the Arduino. The ADC is connected to 3 hall effect sensors attached to the X, Y, and Z axis coils for measuring the respective magnetic fields in real time. Additionally, a PS4 DualShock wireless gaming controller is connected to the Jetson Orin over Bluetooth and used to manually control the generated fields and therefore the microrobots in the workspace. Finally, the Jetson can be connected to a monitor via HDMI to run the software. See Figure \ref{fig:Figure2} for the electrical and system component pipeline of the system. See Figure \ref{fig:Figure3} for a more detailed pin-out representation of the electrical schematic.

\subsubsection{Software}
\begin{figure}
    \centering
    \includegraphics[width=.7\linewidth]{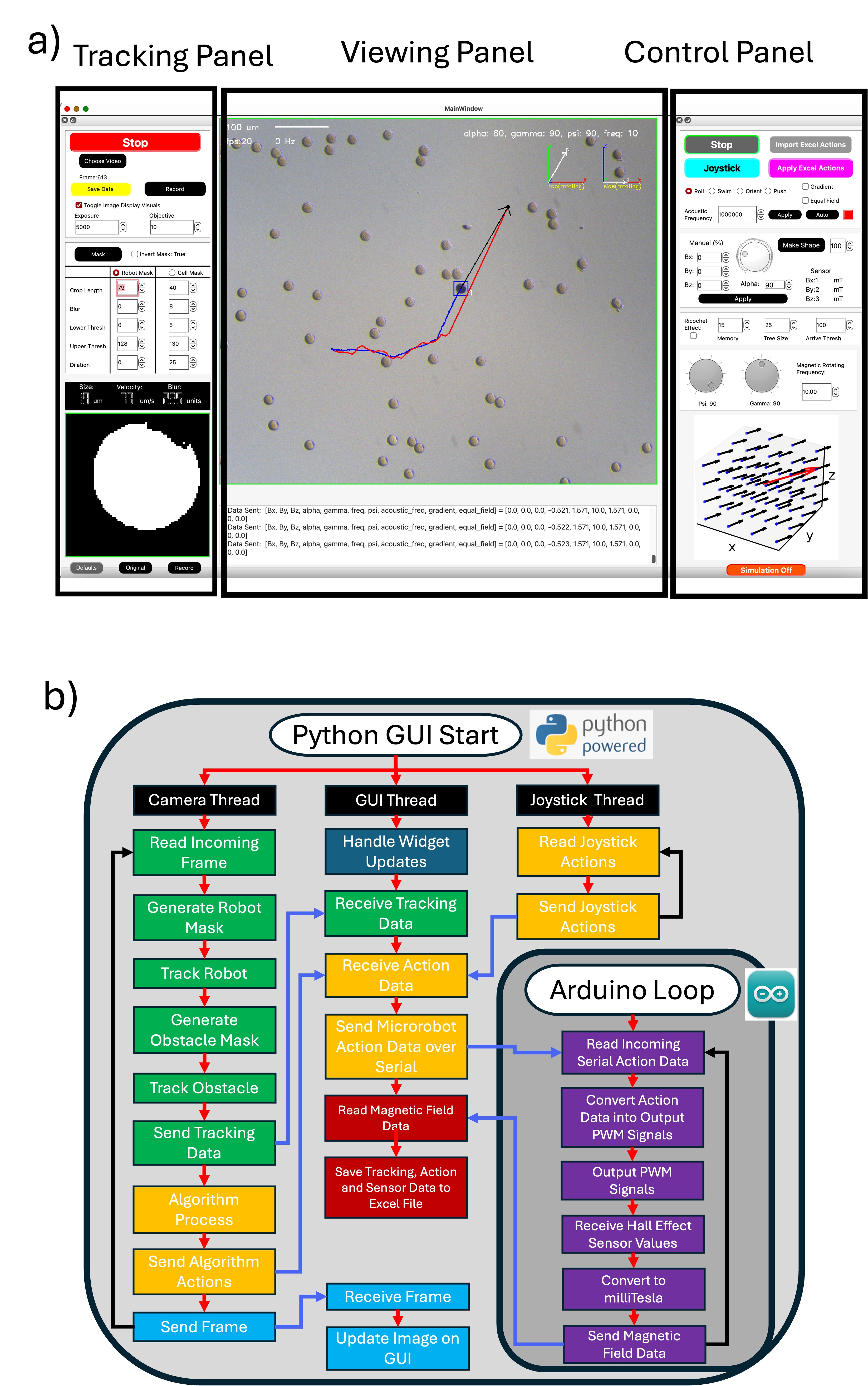}
    \caption{a) Front-end software architecture developed using the PyQt5 library in Python, with tracking, viewing, and control tabs highlighted b)  High-level back-end software architecture flowchart illustrating the main Python GUI processes and corresponding Arduino control loop functionality. Arduino and Python logos included to indicate software/hardware used. Logos are trademarks of their respective owners.}
    \label{fig:Figure4}
\end{figure}

A user-friendly control application graphical user interface was designed in order to easily interact with the camera and adjust control signals. The software is written in Python and is initially built using the QT Designer toolbox from the PyQt5 library which enables simple drag and drop placement of common GUI widgets like buttons and input boxes \cite{pyqt5}. The software front-end architecture can be split into a tracking panel (left), viewing panel (middle), and control panel(right) (see Figure \ref{fig:Figure4}a). Additionally, Figure \ref{fig:Figure4}b outlines the high-level, back-end architecture where black arrows indicate loops within a thread, blue arrows indicate data transmission between threads, and red arrows represent the sequential execution of key functions within individual threads.

The left-most panel of the software consists of widgets and input boxes related to the camera feedback of the system. This panel handles everything related to the incoming camera feed and therefore the tracking and detection of a microrobot in real time.  This includes tuneable parameters that convert the incoming color camera frame into a black and white mask that enables detection and tracking of the microrobot. These tunable parameters include the crop length of the bounding box surrounding the object, image blur, dilation values, and lower/upper mask threshold values. These can all be adjusted in real time to accurately track a wide variety of microrobots. The tracking algorithm resembles a centroid-based object tracking method using binary mask segmentation and center-of-mass localization.  In brief, the tracking algorithm works by first clicking on a location on the viewing panel with a computer mouse. This location should be near the desired microrobot. A bounding square box is then calculated surrounding the location (pixel coordinate) that was selected. Then, a black and white mask is calculated that aims to contrast the microrobot against the background. A successful mask will highlight the microrobot in white and the background in black. The algorithm will then calculate the center of mass of the black and white cropped mask within the bounding box. The calculation of the center of mass results in a pixel coordinate (x,y) describing the position of the microrobot within the bounding box. Then, a new bounding box is calculated centered around this position coordinate. As the microrobot moves around in the workspace, the center of mass is continuously recalculated which generates a new bounding box. This enables real time tracking of an arbitrary microrobot, as long as the black and white mask sufficiently highlights the microrobot. Velocities are calculated using a finite difference estimate over a 15 frame temporal window. The window of 15 frames was chosen to balance noise and resolution. Information such as current mask parameters, position, velocity, and size of the microrobot at each camera frame can be recorded and saved to an excel file for further analysis.  This can be done by pressing on the \textit{\textbf{Save Data}} button to save the raw data only, or \textit{\textbf{Record}} button for recording live videos along with the excel tracking data. This panel also has adjustable variables for changing the exposure of the camera, and an input box for changing the objective magnification to ensure an accurate pixel to metric conversion factor. Finally, at the very bottom of the \textit{Tracking Panel} is a small viewing window of the most recently selected microrobot. The window provides feedback to the user regarding the current black-and-white mask. This allows the user to see in real time how adjusting the tracking parameters affects the mask and therefore the tracking of the microrobot.

The right panel contains everything related to controlling the magnetic field and acoustic frequency applied to the workspace. This includes connecting to a PS4 gaming controller, as well as an automated control button that will apply custom closed loop control algorithms using the microrobot's position and velocity data mentioned above. Additionally, there are buttons for importing an excel actions file which allow for predetermined magnetic or acoustic frequencies to be applied to the workspace. Furthermore, there is a subsection for adjusting the acoustic frequency, applying the acoustic frequency, and an automatic button that sweeps the frequencies in real time to find the resonant actuation frequency of an acoustic bubbled-propelled microrobot. These acoustic frequency signals are applied to the piezoelectric transducer from the AD9850 signal Generator Module. The control panel also contains input boxes for adjusting the magnetic field applied in the x, y, and z directions. A user can switch between a swimming rotating magnetic field, a rolling rotating magnetic field, an orienting uniform field, and a gradient field. These magnetic field output or action parameters can be saved to the same excel file with the tracking information described from the \textit{Tracking Panel}. Finally, a simulation of the applied magnetic vector field direction is also viewable at the bottom right of the control panel as seen in Figure \ref{fig:Figure4}a. 

Finally, the middle panel contains the live image feed from a microscope camera with dimensions of 2048 pixels in the y direction, 2448 pixels in the x direction, and a frame rate set at 24 fps. The frame rate of the camera, a scale bar, and applied acoustic frequency are displayed in the top left of the image. The magnetic field output direction and type (uniform, gradient, or rotating), and current field directions and frequency of rotation ($\alpha$, $\gamma$, and $f$) are displayed in the top right.  This allows a user to analyze how the magnetic or acoustic field affects the microrobot in real time at each frame of the recording. Analyses such as mean square displacement graphs and speed vs frequency graphs can be easily generated from this data.  Furthermore, a user can click on microrobots on the screen to track, manually draw paths for the microrobot to follow, create automated paths that avoid nearby obstacles using path-planning algorithms, and zoom in and out on selected features.  Below the image feedback window is an output log section that displays the currently applied actuation signals that are being sent from the computer to the Arduino.

It should be noted that the software runs on Linux, macOS, and Windows operating systems, as well as on both desktop or laptop computers.  The only requirement is 2 USB ports for the camera and Arduino to connect to. This feature further supports the system's robustness and versatility.

\subsection{Magnetic Field Generation}
\begin{figure}
    \centering
    \includegraphics[width=\linewidth]{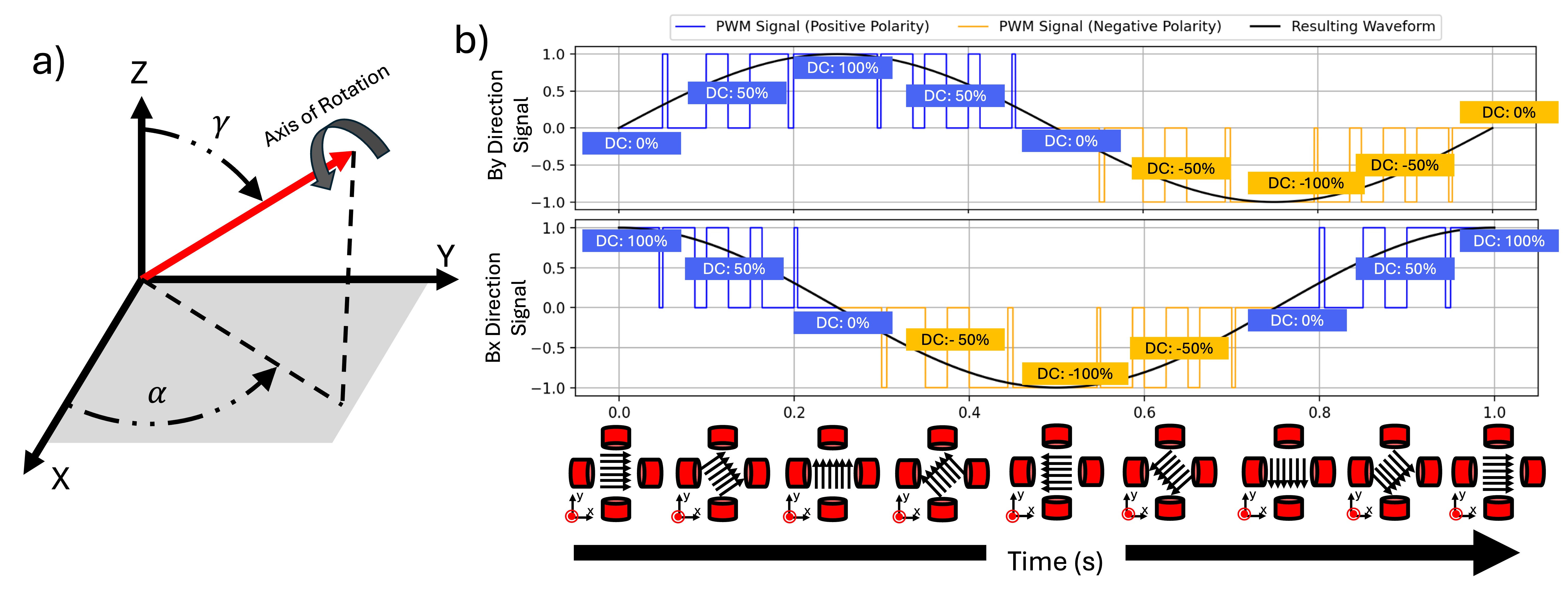}
    \caption{a) Illustration of the spherical coordinate system used to define the axis of rotation of the rotating magnetic field, as described by equations (1–3). The axis orientation is controlled using the angles $\alpha$ and $\gamma$ parameters. Typically $\gamma = 90^{\circ}$ for 2D rolling motion. b) Schematic illustrating how a rotating magnetic field is synthesized using pulse-width modulation (PWM). The two plots show the resulting waveform formed by modulating the duty cycle (DC) of PWM signals over time. For the positive portion of the waveform (black sinusoid), the PWM signal is applied while the H-bridge is set to positive polarity (shown in blue). For the negative portion, the PWM is applied with the H-bridge in negative polarity (orange). The PWM signals effectively approximate the target sinusoidal waveform by varying the duty cycle from 0$\%$ to ±100$\%$. In this example, two sinusoidal waveforms are generated: the top signal is applied to the y-axis coils and the bottom to the x-axis coils, while the z-axis field is set to zero. This configuration creates a 1 Hz rotating magnetic field in the XY plane, as illustrated at the bottom. This is equivalent to setting $\gamma = 0^{\circ}$ using equations (1-3).}
    \label{fig:Figure5}
\end{figure}

The magnetic field action variables described above are sent to the Arduino using the PySerialTransfer python library. The Arduino then converts these variables into magnetic field vectors. Then these vectors are applied to the system using PWM duty cycle modulation \cite{Sun2012}. By varying the duty cycle of the PWM signal we can control the intensity of the magnetic flux density. To generate a uniform field in the x-direction for example, we actuate the right and left configured coils simultaneously. This generates a pseudo uniform field at the center of the workspace. To generate a gradient magnetic field, we only actuate one coil at a time. For example,  to generate a gradient field in the positive x-direction, we actuate the right coil. Equations (\ref{Bx_roll}), (\ref{By_roll}), and (\ref{Bz_roll}) define three software-generated sinusoidal waves applied to each axis of the coil system which generate a rotating magnetic field.

\begin{equation}\label{Bx_roll}
    B_x = -\cos(\gamma) \cos(\alpha) \cos(2 \pi f t) + \sin(\alpha) \sin( 2 \pi f t)
\end{equation}
\begin{equation}\label{By_roll}
    B_y = -\cos(\gamma) \sin(\alpha) \cos(2 \pi f t) - \cos(\alpha) \sin(2 \pi f  t)
\end{equation}
\begin{equation}\label{Bz_roll}
    B_z = \sin(\gamma) \cos(2 \pi f t)
\end{equation}

 where $0 < \alpha < 360^{\circ}$, $0 < \gamma < 180^{\circ}$, and $0 < f < 250$\,Hz. See Figure \ref{fig:Figure5}a for a schematic representation of how these variables adjust the axis of the rotating field. The azimuthal angle ($\alpha$), polar angle ($\gamma$) and frequency $f$ of the rotating field can be adjusted in the software. 

These sinusoidal waveforms are generated artificially through PWM duty cycle modulation in the Arduino "void loop" function and sampled through time by letting $t = micros()/1e6$. This is done by calculating the value of $B_x$, $B_y$, and $B_z$ at each loop of the Arduino program. The amplitude of the sine waves are normalized between -1 and 1. These values are then applied using Arduino's AnalogWrite() function to modulate the PWM duty cycle applied to each H-Bridge driver which therefore adjusts the field strength in time. Figure \ref{fig:Figure5}b conceptually illustrates the waveform generation of a 1 Hz rotating magnetic field in the XY plane. This waveform would spin a magnetic microrobot at 1 Hz in the counterclockwise direction. The blue and orange signals illustrate how varying the duty cycle of the PWM signal can construct an arbitrary waveform. When the PWM signal is blue, we are generating a magnetic field in the positive direction. When the PWM signal turns orange we are generating a magnetic field in the negative direction. This is done by switching the polarity of current applied to the H-Bridge drivers when the amplitude of the waveforms changes sign. Refer to the Arduino code provided in the Supplemental Information for more details.  Additionally, Figure \ref{fig:Figure5}b also illustrates how a 1 Hz rotating magnetic field signal can be used to spin a magnetic microrobot.

We found that the Arduino program loops at a rate of 500 times each second. This, along with the PWM frequency rate, sets an upper limit to the maximum practical sinusoidal frequency that can be applied. The Arduino Mega PWM frequency was increased from the default 490\,Hz to 31\,kHz to maximize the potential artificial sine wave resolution that could be sampled.  At a rotating magnetic field frequency $f = 167\,$Hz, the field output will update every 1/3 of the sinusoidal period.  Essentially, the number of data points that make up the sinusoidal signal is given by $500/f$. For example, at $f = 1\,$Hz, the sinusoidal field will consist of 500 data points. At $f = 10\,$Hz, the sinusoidal field will consist of 50 data points. The minimum number of discrete points within one period that can produce a rolling motion of a ferromagnetic microrobot is just over 2, since at 2 or fewer points aliasing will occur and the field vectors will start to move backward. Therefore, the maximum usable rolling frequency that the system can produce (the Nyquist rate) is approximately 250\,Hz, although lower frequencies may be desired to maintain a truer sinusoidal shape of the magnetic field and therefore smoother rotational motion of the microrobot.

\begin{figure}
    \centering
    \includegraphics[width=1\textwidth]{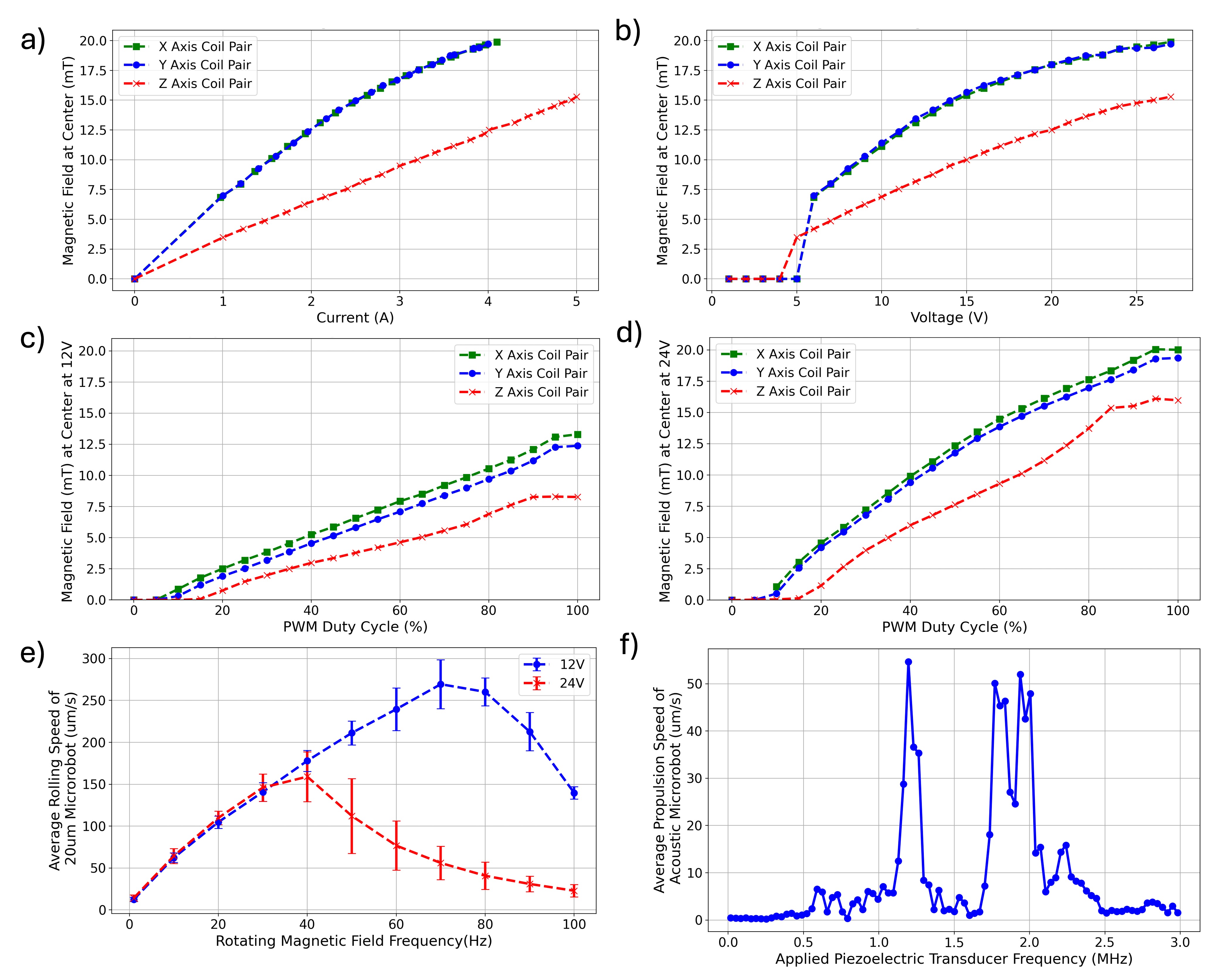}
    \caption{a) Magnetic field (mT) for the x, y, and z coil pairs as a function of current (A). b) Magnetic field (mT) for the x, y, and z coil pairs at set power supply voltages from 0-27V at 100\% duty cycle. c) Magnetic field (mT) for the x, y, and z coil pairs at PWM duty cycles from 0-100\% at 12V supply voltage. d) Magnetic field (mT) for the x, y, and z coil pairs at PWM duty cycles from 0-100\% at 24V supply voltage. e) Average velocity in $\mu$m/s of 20\,$\mu$m rolling microrobot coated in 250\,nm of Nickel at 12V and 24V. f) Average velocity of 3\,$\mu$m cup shaped microrobot at acoustic frequencies ranging from 0\,Hz to 3\,MHz.}
    \label{fig:Figure6}
\end{figure}

\subsection{Magnetic Field Characterization}
The magnetic field at the center of the workspace for the x, y, and z coil pairs was measured using a TD8620 handheld teslameter. This tesla-meter was used to calibrate each hall effect sensor as well. The built-in power supply enables the manual adjustment of the supply voltage from 0V to 27V. Therefore, the intensity of the magnetic field strength and amplitude of the rotating magnetic field can also be increased or decreased.

Figure \ref{fig:Figure6} shows the results of the measurement experiments. It was found that the magnetic field generated by the pair of X coils was comparable to the magnetic field of the Y coil pair. Interestingly, the activation voltage of the H-bridge drivers needed to draw a current load from the systems external power supply was 6\,V for the X and Y coils and 5\,V for the Z Helmholtz coil. At these lowest voltage settings, the X coil was able to draw 1.0\,A generating a field of 6.9\,mT, the Y coil was able to draw 1.0\,A generating a field of 7.0\,mT, and the Z coil was able to draw 1.0 A generating a field of 3.5\,mT. Similarly, as we increase the set voltage of the external power supply in 1\,V increments, we see a relatively linear increase in magnetic field at the center of the workspace \ref{fig:Figure6}b. However, the linearity for the x and y coils decreases after approximately 13 V likely due to heating of the coils and therefore increased resistance in the coils.

The maximum field each coil pair can generate at 27\,V at the center of the workspace is 19.9\,mT in the x-direction, 19.7\,mT in the y-direction, and 15.3\,mT in the z-direction. It was found that the magnetic field generated by the pair of Z coils was more linear in shape than the magnetic field generated by the pair of X and Y coils.  

The voltage of the system is typically set at 12\,V. This voltage provides a sufficient magnetic field to the workspace, while also preventing the coils from overheating. Due to the system operating on PWM (Pulse Width Modulation), the magnetic field strength can be further regulated by varying the duty cycle in software. Figure \ref{fig:Figure6}c shows the magnetic field at the center of the workspace as a function of the PWM duty cycle applied to the system at 12V supply voltage. It is observed that the field increases linearly from 0 mT to approximately 13 mT, 12.5 mT, and 8.3 mT at 100\% duty cycle for the x, y, and z coils respectively. Similarly, Figure \ref{fig:Figure6}d shows the magnetic field as a function of duty cycle at 24V. The field increases from 0 mT to approximately 20 mT, 19.4 mT and 16 mT at 24V for the x, y and z coils respectively. As a result, it can be seen that the system can generate a significant range of magnetic field strengths depending on the application.  These fields are sufficient for actuating a wide variety of magnetic microrobots.

Additionally, the rotating magnetic field of the system was evaluated by measuring a microrobots speed as a function of the rotating frequency as seen in Figure \ref{fig:Figure6}e. A 20\,$\mu$m silica microsphere coated in 250\,nm of Nickel was used as the testing agent. Experiments were done at a voltage setting of both 12\,V and 24\,V to compare the microrobots rolling efficiency. Starting at 1\,Hz the rotating magnetic field frequency was increased in 10\,Hz increments until reaching a maximum of 100\,Hz. At both 12\,V and 24\,V, the microrobots speed was approximately the same until reaching 40\,Hz. At this frequency, a step-out frequency was observed at 12\,V applied to the system. This resulted in a decrease in velocity despite an increase in frequency. The step-out frequency at which magnetic microrobots lose synchronization are affected by several factors, including their magnetic properties, shapes, the strength of the applied field, and the viscosity of the environment \cite{ye2019hydrophobicity}, \cite{WANG2024165}.  The maximum velocity of the microrobot at 12\,V is 160\,$\mu$m/s with the step-out frequency of 40\,Hz. However, at 24\,V, due to the increase in power to the system, and thus an increase in magnetic field, the step-out frequency for the same microrobot was found to be much higher at around 70\,Hz. As a result, he maximum velocity for the same microrobot at 24\,V is 270\,$\mu$m/s.

\begin{figure}
    \centering
     \includegraphics[width=.8\linewidth]{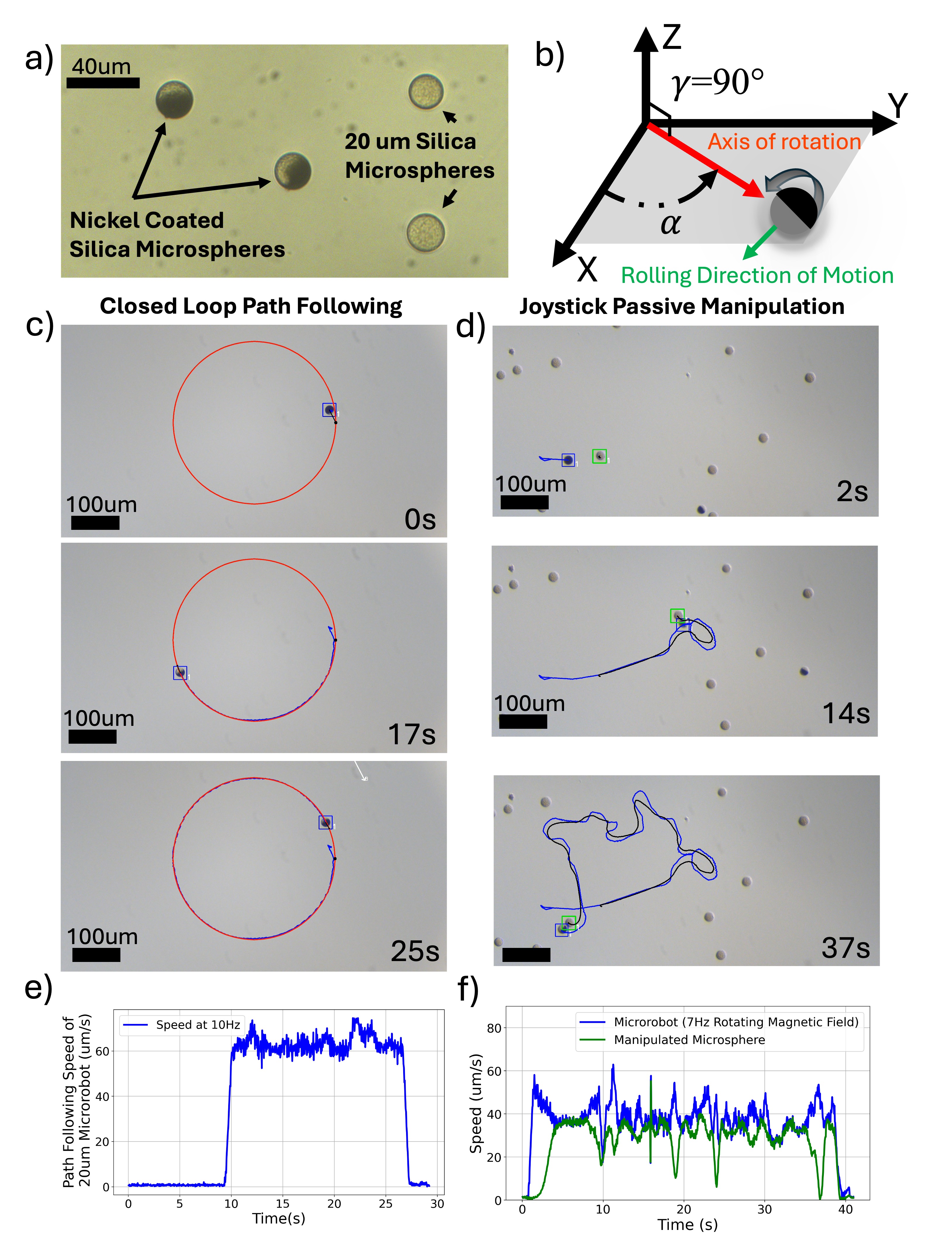}
    \caption{a) Optical microscopy images of magnetic rolling microrobots composed of of 20\,$\mu$m silica microspheres coated in 250\,nm of Ni. b) Schematic representation demonstrating how the axis of rotation can be adjusted to roll a microrobot forward.  20\,$\mu$m rolling magnetic microrobot following a circle trajectory using closed loop control algorithm \ref{alg:algo1}. d) Passive Manipulation of a 20\,$\mu$m tracer particle using a 20\,$\mu$m rolling magnetic microrobot actuated at a rotating magnetic field of 7\,Hz and 12\,V. e) Live speed data over time of the rolling microrobot as it follows the red circular path. f) Live speed data of the microrobot and passive sphere as the microrobot pushes it around the environment.}
    \label{fig:Figure7}
\end{figure}

\subsection{Experimental Validation}
\subsubsection{Closed Loop Path Following of a Rolling Magnetic Microrobot}
First, a simple error minimizing path following algorithm was programmed into the software (see Algorithm \ref{alg:algo1}). The algorithm aims to minimize the distance between a microrobots position $p_x, p_y$ and a target coordinate $(x_i, y_i)$. This is achieved by calculating an angle that points in the direction of the target coordinate. This angle $\alpha$ is determined using $\arctan{\bigg(\dfrac{ (y_i-p_y)}{(x_i - p_x)}\bigg)}$ and is constantly updated at the same rate as the camera frame rate. If the microrobot comes within a certain threshold distance of the target coordinate, the algorithm will move on to the next coordinate in the trajectory array. This allows the microrobot to follow a path consisting of a $N$ nodes or coordinates.

\begin{algorithm}
\caption{Closed Loop Path Following for Rolling Microrobots}\label{alg:algo1}
\begin{algorithmic}[1]
\State \textbf{Input:} Trajectory array of $n$ nodes with positions $(x, y)$
\State \textbf{Output:} Magnetic Field Vector $\mathbf{B} \leftarrow [B_x, B_y, B_z]$
\Statex
\State \textit{Initialization:}
\State $trajectory \gets \begin{bmatrix}
[x_0, y_0], [x_1, y_1], \dots, [x_N, y_N]
\end{bmatrix}^\top$
\State $i \gets 0$
\State $f \gets 10$ Hz
\State $\gamma \gets 90^\circ$
\While{$i < N$}
    \State $\mathbf{p} \gets [p_x, p_y]$
    \State $\mathbf{node} \gets [x_i, y_i]$
    \State $error \gets \sqrt{(x_i - p_x)^2 + (y_i - p_y)^2}$
    \State $\alpha \gets \arctan\left(\dfrac{y_i - p_y}{x_i - p_x}\right)$
    \If{$error < \text{threshold}$}
        \State $i \gets i + 1$
    \Else
        \State $B_x \gets \cos(\gamma)\cos(\alpha)\cos(2 \pi f t) + \sin(\alpha)\sin(2 \pi f t)$
        \State $B_y \gets -\cos(\gamma)\sin(\alpha)\cos(2 \pi f t) + \cos(\alpha)\sin(2 \pi f t)$
        \State $B_z \gets \sin(\gamma)\cos(2 \pi f t)$
        \State \textbf{Apply} $\mathbf{B}$
    \EndIf
\EndWhile
\end{algorithmic}
\end{algorithm}

Figure \ref{fig:Figure7}a illustrates 20\,$\mu$m silica microspheres coated in 250\,nm of Ni and uncoated passive silica microspheres. This microrobot is actuated by a rotating magnetic field. Fig \ref{fig:Figure7}b illustrates how the axis of the rotating magnetic field can be adjusted in any direction to induce rolling motion of the microrobot. Figure \ref{fig:Figure7}c shows the microrobot autonomously following a predefined circular path using Algorithm \ref{alg:algo1}. The red line indicates the desired trajectory which consists of 100 nodes or target coordinates. The blue line indicates the actual robots trajectory. The goal of the algorithm is to minimize the error between the robots current position and the next node in the trajectory array as described above. The frequency of the rotating magnetic field was set to 10\,Hz and the external power supply was set to 12\,V. The microrobot moved at an average velocity of 63.2\,$\mu$m/s (See Figure \ref{fig:Figure7}e). The microrobot took 25 seconds to complete the path and the error between the actual and desired path is minimal as seen in Figure \ref{fig:Figure7}c and Supplemental Video 1.

\subsubsection{Manual Joystick-Controlled Rolling Microrobotic Passive Microsphere Manipulation}
Next, a manual passive particle manipulation experiment was performed using a PS4 dual-shock gaming joystick. Figure \ref{fig:Figure7}d shows snapshots of the 20\,$\mu$m Nickel coated microrobot actively pushing a passive 20\,$\mu$m silica particle while avoiding nearby passive particles. A custom button mapping is used to assign particular actuation variables to the buttons and joysticks of the controller. The right joystick of the controller controls the $0 < \alpha < 360^{\circ}$ variable which therefore determines the heading direction of a rolling microrobot. The left joystick controls the field direction and can be used to orient a magnetic microrobot if the microrobot has an inherent self-propulsion mechanism. The left and right triggers of the controller control the negative and positive magnetic field intensity in the Z direction respectively. The square button on the controller switches $\gamma$ to $180^{\circ}$ resulting in a spinning motion of the microrobot in the clockwise direction. The circle button switches $\gamma$ to $90^{\circ}$ resulting in a spinning motion of the microrobot in the counterclockwise direction. These quick buttons allow for increased control over the behavior of the microrobot when passively manipulating micro objects. In this experiment, the microrobot is tracked using a blue bounding box and blue tracking line. The object to be manipulated is tracked using a green bounding box and a black tracking line. This experiment not only highlights the precision of the system in actively manipulating a passive micro-particle using a gaming controller, but also in the robustness of the custom embedded tracking and detection algorithm. Over the course of approximately 1 minute both the microrobot and manipulated object can be tracked effectively in real time without any tracking malfunction.  The efficiency in tracking enables more sophisticated data driven micro-manipulation algorithms. See Supplemental Video 2 for more details. Figure \ref{fig:Figure7}f shows the velocity of both the microrobot and the passive particle over the 40 second experiment.

 \begin{figure}
    \centering
    \includegraphics[width=\linewidth]{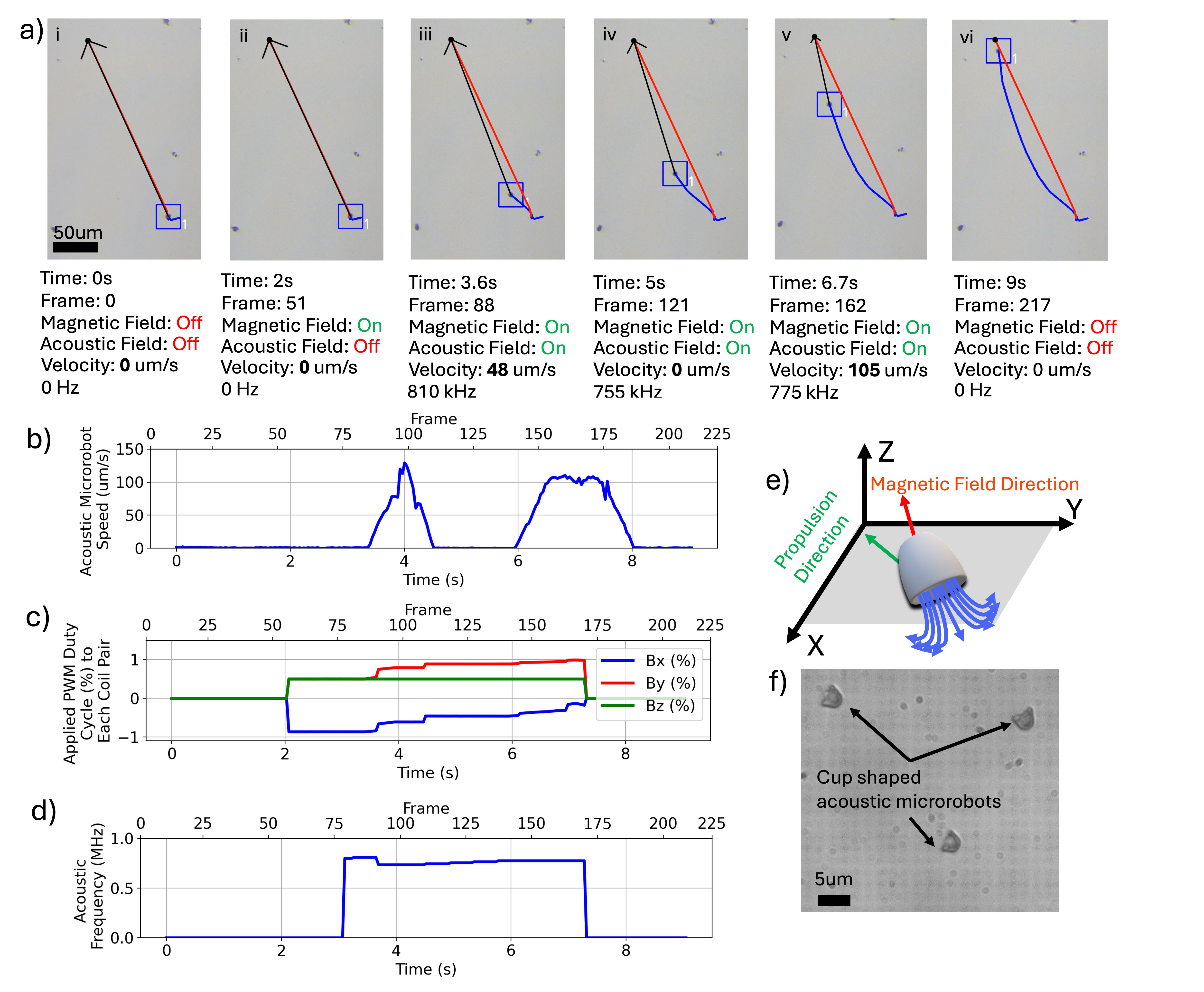}
    \caption{Magnetic and acoustic dual actuation closed loop targeting experiment. a) Snapshots of a 3 um cup shaped microrobot autonomously being guided towards a target location (black arrow). b) The resulting linear velocity of the cup-shaped microrobot over time during the experiment. c) The applied PWM duty cycle applied to the x,y and z axis of the coil system at each frame of the experiment. Negative duty cycles indicate the field was applied in the negative direction. Experiments were conducted at 12V. d) Applied acoustic frequency to the piezo electric transducer over time during this experiment. e) Motion schematic of the acoustic microrobot. f) Optical microscopy images of the cup shaped microrobots under a 50x objective. }
    \label{fig:Figure8}
\end{figure}

\subsubsection{Closed Loop Magnetic and Acoustic Dual Actuation}
Finally, a closed loop navigation experiment was performed to validate the simultaneous magnetic and acoustic actuation functionality of the system. A piezoelectric ring transducer measuring 18\,mm by 12\,mm by 1.2\,mm from StemInc (SMR1812T12R412WL) was used. A standard glass microscope slide measuring 75\,mm by 25\,mm was cut in half and the transducer was glued to the bottom using UV curable glue. The transducer was connected to the system using a 2 pin terminal near the top of the coil stand (See Fig \ref{fig:Figure1}c). Three\,$\mu$m in size cup-shaped acoustic microrobots from \cite{mcneill2020wafer} were used as the microrobotic agent. Figure \ref{fig:Figure8}f shows optical microscopy images of microrobots. The microrobots were scratched from the surface of a silicon wafer using a pipette tip with water and added to the transducer substrate. The propulsion principle is illustrated in Figure \ref{fig:Figure8}d. Figure \ref{fig:Figure6}f shows the speed of the cup shaped robot as the acoustic frequency is swept from 0 MHz to 3 MHz. It is observed that the microrobot's velocity peaks around 1.2 MHz, 1.8 MHz and 2 MHz. These peaks and drops in velocity are potentially the result of multiple resonant modes of the microrobot especially as it moves through the liquid medium. Additionally, the velocity of the microrobot also depends on the microrobot's orientation. Therefore, its orientation changes as it propels which can result in unpredictable motion behavior. However, further work is required to fully understand and characterize these frequency dependent dynamics.  Figure \ref{fig:Figure8} and Supplemental Video 3 illustrates the results of the closed loop autonomous navigation experiment. 

\begin{algorithm}
\caption{Acoustic Microrobot Automated Control Algorithm}
\label{alg:algo2}
\begin{algorithmic}[1]

\State \textbf{Input:} microrobots Velocity Magnitude $v_{mag}$
\State \textbf{Output:} Acoustic Wave Generator Frequency $f_{current}$

\State \textbf{Initialization:}
\State Define $f_{\min}$
\State Define $f_{\max}$
\State $f_{current} \gets f_{\min}$
\State Define $v_{\min}$
\State Define $v_{\max}$
\State $increment \gets 0.1$ MHz
\State $f_{optimal} \gets$ None
\State $n \gets 0$

\While{True}
    \If{$v_{mag} < v_{\min}$}
        \If{$f_{current} < f_{\max}$}
            \If{$n \bmod 10 = 0$}
                \State $f_{current} \gets f_{current} + increment$
            \EndIf
        \ElsIf{$f_{current} \ge f_{\max}$}
            \State $increment \gets increment / 2$
            \State $f_{current} \gets f_{\min}$
        \EndIf
    \ElsIf{$v_{\min} < v_{mag} < v_{\max}$}
        \State $f_{optimal} \gets f_{current}$
        \State $increment \gets increment / 10$
    \ElsIf{$v_{mag} > v_{\max}$}
        \If{$n \bmod 20 = 0$ \textbf{and} $f_{current} > f_{\min}$}
            \State $f_{current} \gets f_{current} - 0.075$ MHz
        \EndIf
    \EndIf
    \State $n \gets n + 1$
\EndWhile

\end{algorithmic}
\end{algorithm}

A closed loop orientation control algorithm is use to direct the microrobot toward a target coordinate. It does this by finding the rotation matrix between the microrobots current direction of motion and the applied magnetic field direction. Once a target coordinate is defined it finds the vector from the microrobots current position and this target vector. It then applies a magnetic field in the direction of this target vector using this rotation matrix. It does this continuously until the microrobot reaches the target coordinate. The algorithm is described in more detail in \cite{sokolich2023automated}. To get the microrobot moving in the first place, an algorithm highlighted in \cite{cerenCAMP} and shown below in \ref{alg:algo2} was implemented into the system to regulate the microrobots' propulsion velocity. 

Figure \ref{fig:Figure8}a illustrates snapshots from the experiment with details of the time, frame number, magnetic field status, acoustic field status, velocity of the microrobot, and applied acoustic frequency. In \ref{fig:Figure8}a(i), at time 0s and frame 0, both the magnetic field and acoustic field are off. Therefore, the velocity of the microrobot is zero. In (ii) the magnetic field is turn on at approximately 2s at frame 51. This begins the orientation algorithm from \cite{sokolich2023automated} which orients the microrobot towards a target coordinate indicated by the black arrow. In (iii), \ref{alg:algo2} begins and finds an acoustic frequency of 810 kHz.  The microrobot begins to propel forward and the magnetic field adjusts itself in real time to guide the microrobot in the direction of the black target vector. However, in (iv) the acoustic frequency drops to 755 kHz resulting in the microrobot coming to a stop. The algorithm recognizes the microrobot has zero velocity despite an applied acoustic frequency. As a result, in (v) the acoustic frequency is autonomously increased to 775 kHz which resumes the microrobots propulsion before finally reaching the target vector head in (vi) and concluding the closed loop control sequence. 

Figure \ref{fig:Figure8}b shows a graph of the microrobot's real time velocity over time from this closed loop control sequence.  Figure \ref{fig:Figure8}b depicts a graph of the PWM duty cycle percentage applied to the x,y and z pair of coils over time based on the orientation algorithm. This is proportional to the magnetic field applied in each direction. A negative percentage indicates a field was applied in the negative direction.  During the experiment, the PWM duty cycle in the z direction is set to $50\%$ which allows the microrobot to orient itself slightly upwards. Finally, Figure \ref{fig:Figure8}c depicts a graph of the acoustic frequency applied to the piezoelectric transducer over time. These frequencies are adjusted based on the microrobots velocity from \ref{alg:algo2}.

\section{Conclusion}
This paper presents a versatile and portable platform for data-driven acoustic and magnetic microrobotic experimentation. All system-wide mechanical, electrical, and software components involved in constructing the instrument were described. Then we outline the magnetic field generation method.

Next, we provide magnetic field measurements at various system settings at the center of the workspace. To validate the system’s experimental versatility, we demonstrate several representative microrobotic operations commonly found in the literature. These include autonomous path following of a rolling microrobot and joystick-controlled passive particle manipulation. We also demonstrate vision-based acoustic and magnetic microrobot navigation enabled by the simultaneous generation and control of both fields.

This integrated platform offers a powerful and efficient solution for microrobotic experimentation. A user-friendly control system of this size and functionality could be used in a microrobotic lab-based graduate courses similar to \cite{cappelleri2025design}. Overall, the work represents a significant step toward transitioning microrobotic systems from specialized laboratory setups to broader research, commercial, and educational environments.

\section*{Acknowledgment}
This work was supported by the National Science Foundation under grant GCR 2219101 and the National Health Institute under grant 1R35GM147451. This project was also supported with a grant from the National Institute of General Medical Sciences – NIGMS (5P20GM109021-07) from the National Institutes of Health and the State of Delaware. 

\section*{Author Contributions}
M.S. conceptualized idea, designed mechanical, electrical and software architecture, designed and conducted experiments, and wrote and revised manuscript.  Y.Y. helped create videos. FC.K., S.C. and S.D. helped with writing. 

\bibliographystyle{ieeetr}
\bibliography{bibliography}

\begin{thebibliography}{10}

\bibitem{doi:10.1021/acsnano.3c03723}
J.~G. Lee, R.~R. Raj, N.~B. Day, and C.~W.~I. Shields, ``Microrobots for biomedicine: Unsolved challenges and opportunities for translation,'' {\em ACS Nano}, vol.~17, no.~15, pp.~14196--14204, 2023.
\newblock PMID: 37494584.

\bibitem{li2025inhalable}
Z.~Li, Z.~Guo, F.~Zhang, L.~Sun, H.~Luan, Z.~Fang, J.~L. Dedrick, Y.~Zhang, C.~Tang, A.~Zhu, {\em et~al.}, ``Inhalable biohybrid microrobots: a non-invasive approach for lung treatment,'' {\em Nature Communications}, vol.~16, no.~1, p.~666, 2025.

\bibitem{tran2025robotic}
H.~H. Tran, Z.~Xiang, M.~J. Oh, Y.~Liu, Z.~Ren, C.~Chen, N.~Jaruchotiratanasakul, J.~M. Kikkawa, D.~Lee, H.~Koo, {\em et~al.}, ``Robotic microcapsule assemblies with adaptive mobility for targeted treatment of rugged biological microenvironments,'' {\em ACS nano}, 2025.

\bibitem{chen2022electromagnetic}
R.~Chen and D.~Folio, ``Electromagnetic actuation microrobotic systems,'' {\em Current Robotics Reports}, vol.~3, no.~3, pp.~119--126, 2022.

\bibitem{shen2023magnetically}
H.~Shen, S.~Cai, Z.~Wang, Z.~Ge, and W.~Yang, ``Magnetically driven microrobots: Recent progress and future development,'' {\em Materials \& Design}, vol.~227, p.~111735, 2023.

\bibitem{10440624}
M.~Naik and E.~Diller, ``Object rotation and translation for serial planar acoustic microassembly,'' {\em IEEE/ASME Transactions on Mechatronics}, vol.~29, no.~5, pp.~3799--3809, 2024.

\bibitem{8629946}
O.~Youssefi and E.~Diller, ``Contactless robotic micromanipulation in air using a magneto-acoustic system,'' {\em IEEE Robotics and Automation Letters}, vol.~4, no.~2, pp.~1580--1586, 2019.

\bibitem{sokolich2023modmag}
M.~Sokolich, D.~Rivas, Y.~Yang, M.~Duey, and S.~Das, ``Modmag: A modular magnetic micro-robotic manipulation device,'' {\em MethodsX}, vol.~10, p.~102171, 2023.

\bibitem{foroutan2018sat}
V.~Foroutan, F.~Farzami, D.~Erricolo, R.~Majumdar, and I.~Paprotny, ``Sat-c: An efficient control strategy for assembly of heterogeneous stress-engineered mems microrobots,'' in {\em 2018 IEEE International Conference on Robotics and Automation (ICRA)}, pp.~2700--2707, IEEE, 2018.

\bibitem{5595508}
M.~P. Kummer, J.~J. Abbott, B.~E. Kratochvil, R.~Borer, A.~Sengul, and B.~J. Nelson, ``Octomag: An electromagnetic system for 5-dof wireless micromanipulation,'' {\em IEEE Transactions on Robotics}, vol.~26, no.~6, pp.~1006--1017, 2010.

\bibitem{aghakhani2020acoustically}
A.~Aghakhani, O.~Yasa, P.~Wrede, and M.~Sitti, ``Acoustically powered surface-slipping mobile microrobots,'' {\em Proceedings of the National Academy of Sciences}, vol.~117, no.~7, pp.~3469--3477, 2020.

\bibitem{bertin2015propulsion}
N.~Bertin, T.~A. Spelman, O.~Stephan, L.~Gredy, M.~Bouriau, E.~Lauga, and P.~Marmottant, ``Propulsion of bubble-based acoustic microswimmers,'' {\em Physical Review Applied}, vol.~4, no.~6, p.~064012, 2015.

\bibitem{ren20193d}
L.~Ren, N.~Nama, J.~M. McNeill, F.~Soto, Z.~Yan, W.~Liu, W.~Wang, J.~Wang, and T.~E. Mallouk, ``3d steerable, acoustically powered microswimmers for single-particle manipulation,'' {\em Science advances}, vol.~5, no.~10, p.~eaax3084, 2019.

\bibitem{ahmed2016artificial}
D.~Ahmed, T.~Baasch, B.~Jang, S.~Pane, J.~Dual, and B.~J. Nelson, ``Artificial swimmers propelled by acoustically activated flagella,'' {\em Nano letters}, vol.~16, no.~8, pp.~4968--4974, 2016.

\bibitem{pyqt5}
R.~Computing, ``Pyqt5.'' \url{https://www.riverbankcomputing.com/software/pyqt/}, 2021.
\newblock Version 5.x.

\bibitem{Sun2012}
J.~Sun, {\em Pulse-Width Modulation}, pp.~25--61.
\newblock London: Springer London, 2012.

\bibitem{ye2019hydrophobicity}
C.~Ye, J.~Liu, X.~Wu, B.~Wang, L.~Zhang, Y.~Zheng, and T.~Xu, ``Hydrophobicity influence on swimming performance of magnetically driven miniature helical swimmers,'' {\em Micromachines}, vol.~10, no.~3, p.~175, 2019.

\bibitem{WANG2024165}
Z.~Wang, W.~Li, A.~Klingner, Y.~Pei, S.~Misra, and I.~S. Khalil, ``Magnetic control of soft microrobots near step-out frequency: Characterization and analysis,'' {\em Computational and Structural Biotechnology Journal}, vol.~25, pp.~165--176, 2024.

\bibitem{mcneill2020wafer}
J.~M. McNeill, N.~Nama, J.~M. Braxton, and T.~E. Mallouk, ``Wafer-scale fabrication of micro-to nanoscale bubble swimmers and their fast autonomous propulsion by ultrasound,'' {\em ACS nano}, vol.~14, no.~6, pp.~7520--7528, 2020.

\bibitem{sokolich2023automated}
M.~Sokolich, D.~Rivas, Z.~H. Shah, and S.~Das, ``Automated control of catalytic janus micromotors,'' {\em MRS Advances}, vol.~8, no.~18, pp.~1005--1009, 2023.

\bibitem{cerenCAMP}
F.~C. Kirmizitas, D.~P. Rivas, M.~Sokolich, J.~M. McNeill, A.~Dutta, and S.~Das, ``Camp: Closed-loop acoustic-magnetic propulsion of microrobots for precision cell manipulation,'' {\em Advanced Intelligent Systems}, vol.~n/a, no.~n/a, p.~2500300.

\bibitem{cappelleri2025design}
D.~J. Cappelleri and I.~Gong, ``Design of a lab-based graduate mobile microrobotics course,'' in {\em 2025 International Conference on Manipulation, Automation and Robotics at Small Scales (MARSS)}, pp.~1--6, IEEE, 2025.

\end{thebibliography}

\end{document}